# Temporal-Aware Fusion for Robust Outdoor LiDAR Localization

Minghang Zhu*
Fujian Key Laboratory of Urban Intelligent Sensing and Computing, Xiamen University
Xiamen, China
mihoo@stu.xmu.edu.cn

Zhijing Wang*
Fujian Key Laboratory of Urban Intelligent Sensing and Computing, Xiamen University
Xiamen, China
23020241154448@stu.xmu.edu.cn

Yuxin Guo
Fujian Key Laboratory of Urban Intelligent Sensing and Computing, Xiamen University
Xiamen, China
23020240157665@stu.xmu.edu.cn

Chen Liu
Fujian Key Laboratory of Urban Intelligent Sensing and Computing, Xiamen University
Xiamen, China
23020231154150@stu.xmu.edu.cn

Yongshu Huang
Fujian Key Laboratory of Urban Intelligent Sensing and Computing, Xiamen University
Xiamen, China
23020231154192@stu.xmu.edu.cn

Wen Li
School of Engineering Mathematics and Technology, University of Bristol
Bristol, United Kingdom
liwen777@stu.xmu.edu.cn

Sheng Ao†
Fujian Key Laboratory of Urban Intelligent Sensing and Computing, Xiamen University
Xiamen, China
aosh@xmu.edu.cn

Cheng Wang
Fujian Key Laboratory of Urban Intelligent Sensing and Computing, Xiamen University
Xiamen, China
cwang@xmu.edu.cn

## Abstract

LiDAR relocalization aims to estimate the global 6-DoF pose of a sensor in the environment. However, existing regression-based approaches often encounter limitations in dynamic or ambiguous scenarios, as they typically prioritize single-frame inference, leaving the potential of spatio-temporal consistency across scans not fully explored. In this paper, we propose a **Temp**oral-aware **Loc**alization framework (**TempLoc**) designed to enhance the robustness of outdoor localization by effectively modeling sequential consistency. Specifically, a Global Coordinate Estimation module is first introduced to predict point-wise global coordinates and associated uncertainties for each LiDAR scan. A Prior Coordinate Generation module is then presented to estimate inter-frame point correspondences by the attention mechanism. Lastly, an Uncertainty-Guided Coordinate Fusion module is deployed to integrate both predictions of point correspondence in an end-to-end fashion, yielding a more temporally consistent and accurate global 6-DoF pose. Experimental results on the NCLT and Oxford RobotCar benchmarks show that our TempLoc outperforms state-of-the-art methods by a large margin, demonstrating the effectiveness of temporal-aware correspondence modeling in LiDAR relocalization.

*Both authors contributed equally to this research.
†Corresponding author.

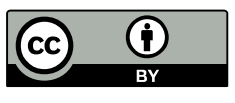

## 1 Introduction

Accurate and robust LiDAR-based relocalization is a fundamental capability for autonomous systems [27, 58, 59] and robotics [36, 42, 48]. Given a pre-built 3D map, the goal of LiDAR-based localization is to estimate the global 6 degree of freedom (DoF) pose of sensor according to the captured real-time scans [2, 3, 35]. The primary difficulty lies in handling an environment with dynamic variations, occlusions, and sensor noise [1, 14, 24, 54].

Most existing LiDAR relocalization methods [7, 17] rely on a matching paradigm, where the input point cloud is aligned against a pre-built global map to estimate the current pose [26, 30]. While highly accurate, these methods require storing and querying large-scale 3D maps, resulting in significant memory and communication overhead [13, 33, 37]. In contrast, recent regression-based methods [44, 45, 57] attempt to directly predict the absolute pose from a single point cloud scan using deep neural networks, bypassing the need for explicit map storage and achieving higher computational efficiency. These regression-based alternatives open up promising directions for deployment on resource-constrained platforms [53].

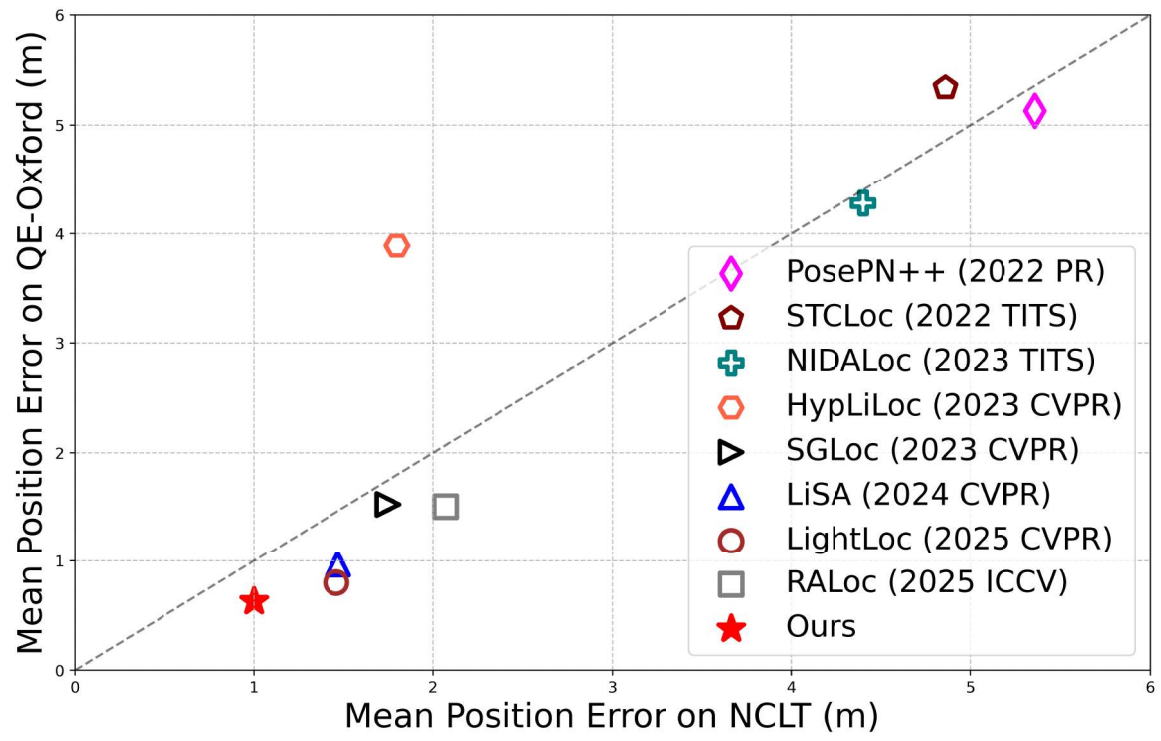

**Figure 1: Mean position error comparisons on NCLT and QE-Oxford dataset. Our method (TempLoc) achieves superior localization accuracy on both datasets.**

Nevertheless, the regression-based methods often struggle in complex or ambiguous environments [44, 51]. This is because single LiDAR scan is prone to measurement noise and observable geometric variations induced by spatial changes or dynamic objects. To alleviate this issue, a handful of work [55, 56] has been proposed by incorporating LiDAR sequences to enhance localization accuracy. However, they only roughly encode spatial-temporal global features, without explicitly modeling the point-level sequential consistency [12], which results in suboptimal localization performance.

In the 2D counterpart, recent advances in temporal relocalization have demonstrated that incorporating frame-to-frame correspondences at the pixel level can significantly improve performance [9, 49, 61]. For instance, KFNet [60] estimates per-pixel scene coordinates and model the temporal transition using optical flow, enabling recursive state estimation through Kalman filtering. While effective for 2D image, these approaches [8, 10, 50] cannot be directly extended to the 3D domain due to the unordered and irregular nature of point clouds in 3D scenes, coupled with challenges such as sparsity, occlusions, and interference from dynamic objects, temporal LiDAR-based localization faces unique difficulties.

This observation raises a critical question: can a similar paradigm be adapted to LiDAR relocalization? Motivated by this, we propose a new LiDAR relocalization framework that jointly performs measurement and prior estimation, integrating them through a learned fusion mechanism. Our method, named TempLoc, consists of three components: (1) the **Global Coordinate Estimation** that predicts per-point global coordinates and uncertainties from a single LiDAR scan, (2) the **Prior Coordinate Generation** that leverages a point cloud registration network to estimate inter-frame point-wise transitions, and (3) the **Uncertainty-Guided Coordinate Fusion** that employs an end-to-end differentiable fusion mechanism to generate accurate and temporally consistent point correspondences. In the end, the global 6-DoF pose is obtained via a RANSAC-based optimization using the refined 3D correspondences.

Benefiting from explicitly modeling the temporal evolution of scene coordinates, our method effectively mitigates the limitations of single-frame regression and significantly enhances localization robustness. As shown in Figure 1, our method achieves state-of-the-art performance on two public available benchmarks. Notably, the localization accuracy of our method outperforms that of the strong baseline LightLoc [21] by nearly 30% on the NCLT [6] dataset. Overall, our contributions are three-fold:

- We propose a temporally-aware LiDAR relocalization framework that incorporates sequential constraints into point correspondence estimation by extending scene coordinate regression to the temporal domain.
- We introduce an uncertainty-aware coordinate estimation module to predicts per-point global coordinates from single LiDAR scan.
- Extensive experiments and ablations demonstrating the effectiveness of our method and providing the intuition for its architectural components.

## 2 Related Work

Map-based methods [19, 39] need to pre-build feature maps and perform localization via retrieval or registration, requiring substantial storage for large-scale outdoor deployments. In contrast, regression-based methods offer a map-free alternative by implicitly encoding maps within network parameters, enabling direct global pose prediction without any additional storage overhead. These regression-based methods can be further categorized into Absolute Pose Regression (APR) and Scene Coordinate Regression (SCR).

### 2.1 Absolute Pose Regression

Absolute pose regression aims to use deep networks to learn and memorize scene information, directly regressing the sensor's global pose in an end-to-end manner. PoseNet [16], a classic solution for APR, applies a modified GoogleNet [38] to regress camera poses. The first proposed APR method for LiDAR is PointLoc [44], which uses PointNet++ [32] and a self-attention module to extract global features. PosePN [57] adopts a universal encoder and memory-aware regression to avoid redundant retraining and improve localization performance. HypLiLoc [43] combines 3D features and 2D spherical projection features to further improve performance. DiffLoc [22] introduces a diffusion model to obtain robust and accurate positioning through iterative denoising. These methods use single-frame data as input, which will results in many outliers.

Since sequential data can provide contextual information and time constraints, VLocNet [40] and its semantic variant VLocNet++ [34] use multi-task optimization strategy to learn shared features, LSG [49] integrates visual odometry constraints, and ViPR [29] integrates APR and relative pose estimation using LSTMs. For filtering and handling specific constraints, CoordiNet [28] eliminates positioning outliers based on Kalman filtering, while LiDAR-specific schemes like STCLoc [56] and NIDALoc [55] apply spatial-temporal regularization and memory modules, respectively. Despite improved accuracy, these sequence-based methods rely on long-term sequence information, causing high computational overhead that hinders real-time performance.

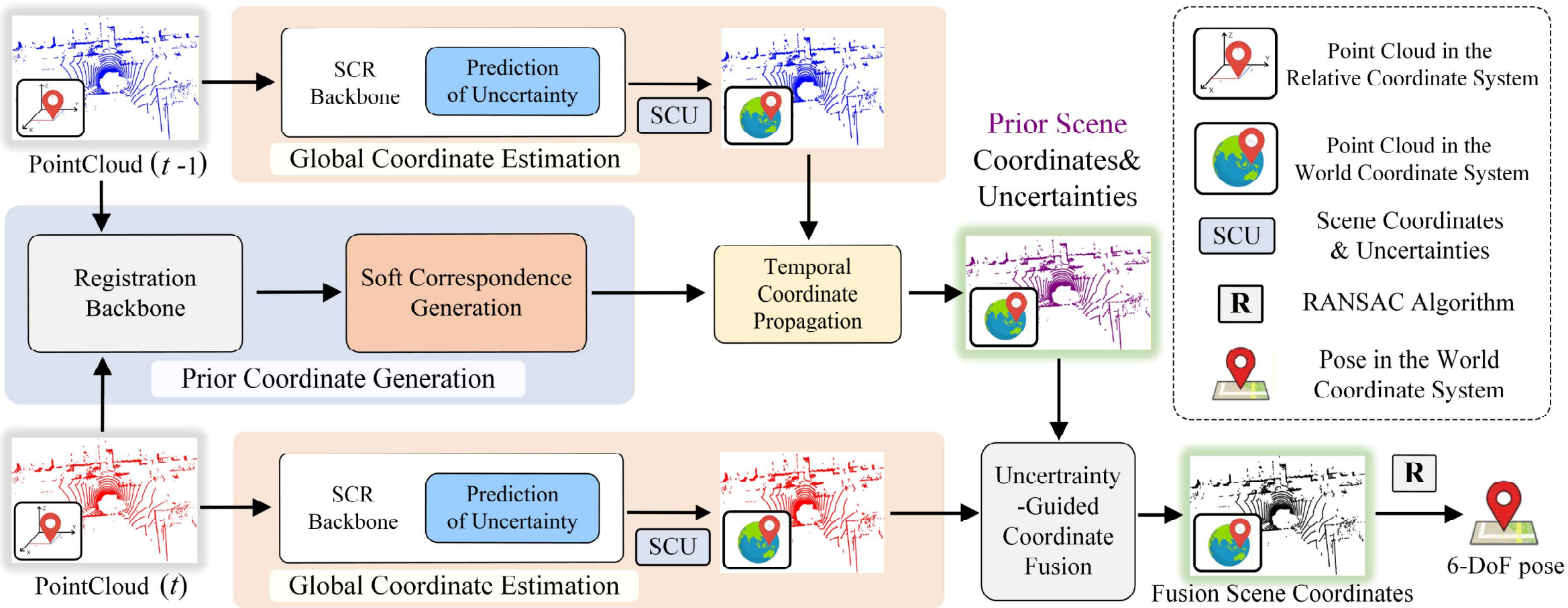

**Figure 2: Overview of the proposed framework. The Global Coordinate Estimation (GCE) module processes sequential point clouds from consecutive frames via a SCR module with uncertainty estimation, producing stable pose estimates. Subsequently, a Prior Coordinate Generation (PCG) module refines the pose estimates using soft correspondences. Then, Uncertainty-Guided Coordinate Fusion (UCF) module produces point clouds in the world coordinate system. Finally, the RANSAC algorithm is employed to solve for the point cloud localization pose.**

## 2.2 Scene Coordinate Regression

Different from APR which directly regresses the pose, SCR regresses the point cloud coordinates in the world coordinate system and relies on RANSAC for pose estimation. SGLoc [23] applied SCR to LiDAR positioning for the first time and significantly improved the positioning performance. LiSA [51] applies diffusion-based knowledge distillation to enable the SCR network to have semantic understanding and focus on more important points. In order to reduce training time, LightLoc [21] proposes a universal encoder that substantially reduces training time without compromising model performance.

However, Scene Coordinate Regression (SCR) often yields numerous outliers in regressed point coordinates. Despite iterative denoising with RANSAC, these outliers can still compromise localization accuracy. To mitigate the impact of such outliers, we integrate uncertainty estimation. Currently, within a filter framework, we fuse temporal information from neighboring frames to achieve superior pose estimation and smoother trajectories.

## 3 Methodology

As shown in Figure 2, our method consists of three key modules: Global Coordinate Estimation (GCE), Prior Coordinate Generation (PCG), and Uncertainty-guided Coordinate Fusion (UCF). The GCE module produces 3D point correspondences for each LiDAR scan between the local and global coordinate systems, along with their associated uncertainty scores (Sec. 3.1). Simultaneously, the PCG module estimates 3D point correspondences between adjacent frames and provides corresponding credibility assessments (Sec. 3.2). Based on this, the UCF module integrates the outputs of the GCE and PCG to obtain refined 3D-3D correspondences between the local and global coordinate systems (Sec. 3.3). Finally, the global 6-DoF pose is computed through RANSAC.

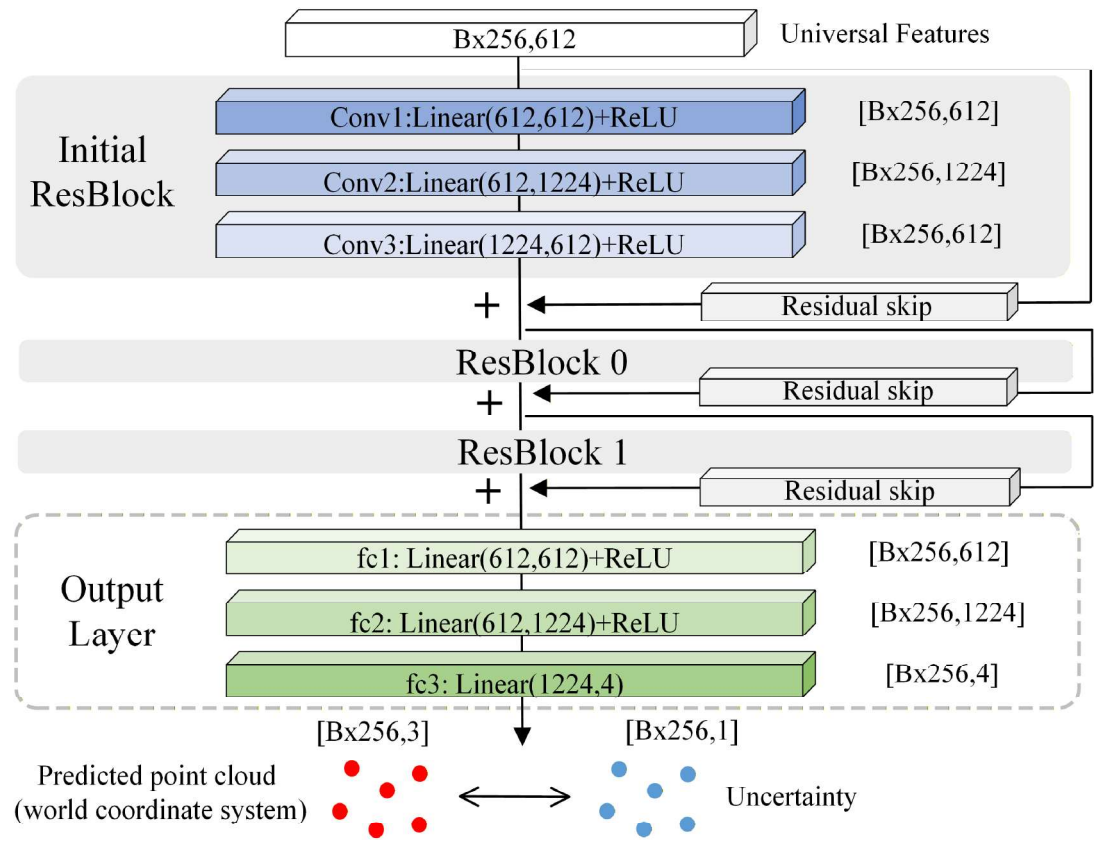

**Figure 3: Global Coordinate Estimation Module. The GCE module takes sparse point cloud features extracted by the backbone network as input. It first extracts deep features through an initial residual block followed by two stacked residual blocks (ResBlock 0 and 1). Finally, three fully connected layers output the predicted point cloud in the world coordinate system along with the uncertainty for each point.**

### 3.1 Global Coordinate Estimation

The SCR method is prone to producing inaccurate coordinate predictions in environments characterized by noise, dynamic disturbances, or insufficient structural information. However, existing LiDAR relocalization methods typically overlook the quality of per-point predictions. To address this, as shown in Fig 3, we incorporate a per-point uncertainty estimation mechanism into the SCR architecture to assess the quality of each predicted global coordinate.

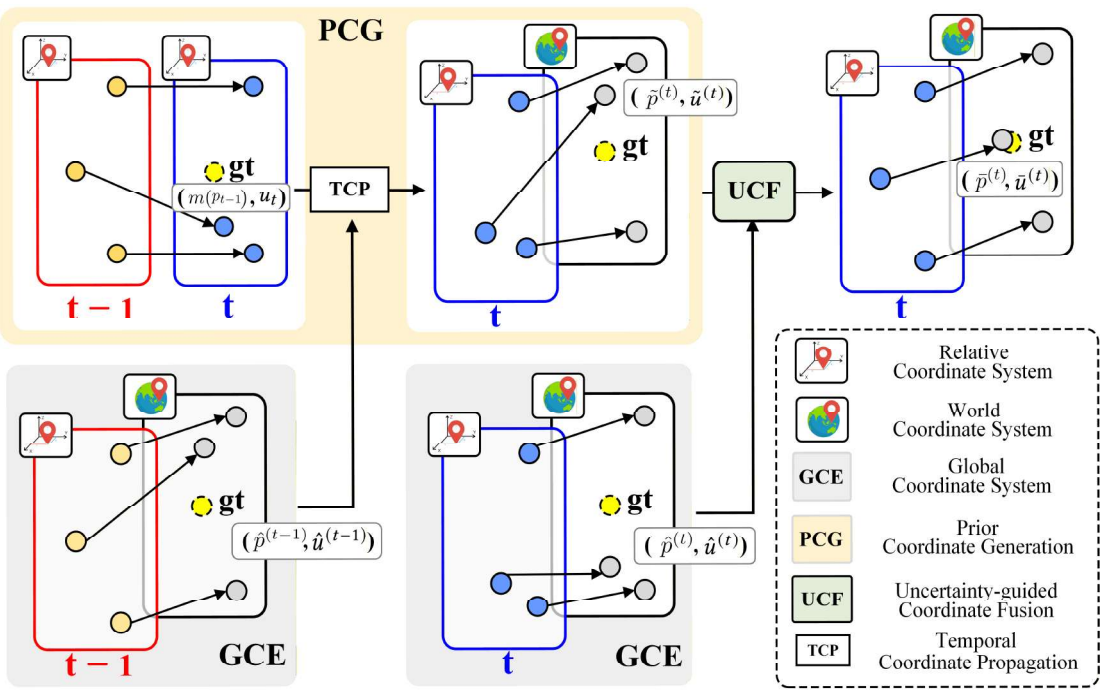

**Figure 4: Illustration of coordinate fusion. The GCE module produces the measurement estimation at $t$ time, while the PCG module provides the prior estimation at the same time. These two estimations are fused by the UCF module to generate more accurate point correspondences between the relative and world coordinate systems.**

***Prediction of uncertainty.*** For an input point cloud $\mathbf{P} \in \mathbb{R}^{N\times3}$, we utilize the LightLoc network [21] to regress scene coordinates $\mathbf{P}_{pred} \in \mathbb{R}^{M\times3}$ in global reference frame. Simultaneously, the per-point uncertainty score $\boldsymbol{u}_{pred} \in \mathbb{R}^{M\times1}$ can be predicted by shared MLPs. Note that, the ground-truth score $u_{gt}$ is defined as:

$$u_i^{gt} = \begin{cases} 0, & \sum_{i=1}^{M} \left\| \boldsymbol{p}_i^{pred} - \boldsymbol{p}_i^{gt} \right\|_1 < \tau, \\ 1, & \text{otherwise.} \end{cases} \tag{1}$$

where $\boldsymbol{p}^{gt}$ represents the ground-truth global coordinate of each point. The threshold $\tau$ is not static and decays as the network training progresses.

This strategy is critical. In early training, a higher $\tau$ permits the network to treat a broad range of predictions as confident, helping it focus on penalizing severe outliers and avoiding the trivial all-high-uncertainty solution. As the network converges and predictions improve, a smaller $\tau$ tightens the criterion, encouraging finer-grained uncertainty estimation.

***Loss Function.*** The total loss is composed of a regression loss and an uncertainty loss. The uncertainty prediction is supervised by a dedicated loss term, $\mathcal{L}_{un}$. We employ the Mean Squared Error loss, which computes the average of the squared differences between the predicted uncertainty $u_i^{pred}$ and the ground truth label $u_i^{gt}$:

$$\mathcal{L}_{un} = \frac{1}{M} \sum_{i=1}^{M} (u_i^{pred} - u_i^{gt})^2. \tag{2}$$

Meanwhile, the primary regression objective, $\mathcal{L}_{reg}$, minimizes the mean distance L1 [11] between the predicted and ground truth coordinates:

$$\mathcal{L}_{reg} = \frac{1}{M} \sum_{i=1}^{M} \left\| \boldsymbol{p}_i^{pred} - \boldsymbol{p}_i^{gt} \right\|_1. \tag{3}$$

The final loss $\mathcal{L}_{GCE}$ is the sum of these two terms:

$$\mathcal{L}_{GCE} = \mathcal{L}_{reg} + \mathcal{L}_{un}. \tag{4}$$

## 3.2 Prior Coordinate Generation

To effectively leverage temporal information, we design a prior coordinate generation module to generate a reliable world coordinate prior for the current frame. This process has two core stages: first, generating a soft correspondence and uncertainty in the current frame for each point in the previous frame using a registration network; second, using these correspondences to assign prior coordinates to points in the current frame through a neighborhood interpolation.

***Soft Correspondence Generation.*** We adopt the PCAM [5] as the backbone network and additionally introduce a self-attention mechanism [41, 62] to enhance point-wise features within each point cloud, thereby generating discriminative feature descriptors.

For a point cloud $\mathbf{P}^{(t)}$ with initial features $\mathbf{F}^{(t)}$, we first compute its self-attention matrix $\mathbf{A}^{(t,t)}$. Each element $(\mathbf{A}^{(t,t)})_{ij}$, representing the attention of point $i$ on point $j$, is computed from the cosine similarity of intra-cloud feature pairs, followed by a Softmax normalization:

$$s_{ij} = \frac{\mathbf{F}_i^{(t)} \cdot (\mathbf{F}_j^{(t)})^T}{||\mathbf{F}_i^{(t)}||_2 \cdot ||\mathbf{F}_j^{(t)}||_2}, \tag{5}$$

$$(\mathbf{A}^{(t,t)})_{ij} = \frac{e^{s_{ij}}}{\sum_{k=1}^{N} e^{s_{ik}}} \quad . \tag{6}$$

After obtaining the context-aware features, we use them to compute multi-level cross-attention from current frame $\mathbf{P}^{(t)}$ to previous frame $\mathbf{P}^{(t-1)}$. We aggregate all $L$ cross-attention matrices $\mathbf{A}_{(l)}^{(t-1,t)}$ through element-wise multiplication to obtain a global attention matrix $\mathbf{A}_{global}^{(t-1,t)}$:

$$\mathbf{A}_{global}^{(t-1,t)} = \mathbf{A}_{(1)}^{(t-1,t)} \odot \mathbf{A}_{(2)}^{(t-1,t)} \odot \cdots \odot \mathbf{A}_{(L)}^{(t-1,t)}. \tag{7}$$

Based on this global attention matrix, we generate a soft correspondence $m(\boldsymbol{p}_i^{(t-1)})$ in the current frame for each point $\boldsymbol{p}_i^{(t-1)}$ from the previous frame. This soft correspondence is a weighted average of all points in the current frame, with weights given by the corresponding row vector of the global attention matrix:

$$m(\boldsymbol{p}_i^{(t-1)}) = \frac{\sum_{j=1}^{N} (\mathbf{A}_{global}^{(t-1,t)})_{ij} \cdot \boldsymbol{p}_j^{(t)}}{\sum_{k=1}^{N} (\mathbf{A}_{global}^{(t-1,t)})_{ik}}, \tag{8}$$

where $\boldsymbol{p}_j^{(t)}$ is the point in current frame cloud $\mathbf{P}^{(t)}$, and $N$ is its number of points. This yields a set of pseudo-correspondence points $\{m(\boldsymbol{p}_i^{(t-1)})\}$, where each point carries the temporal information of its origin point $\boldsymbol{p}_i^{(t-1)}$.

***Temporal Coordinate Propagation.*** The generated soft correspondence point $m(\boldsymbol{p}_i^{(t-1)})$ is a theoretical position and does not necessarily match any of the actual sampled points $\boldsymbol{p}_j^{(t)}$ in the current frame. To assign a prior coordinate to the actual points of the current frame, we use a neighborhood interpolation strategy.

Assume we have predicted the world coordinates of all points in the previous frame, denoted as $\{\hat{\boldsymbol{p}}_i^{(t-1)}\}$. For each actual point $\boldsymbol{p}_j^{(t)}$ in the current frame, we search for its $k$ nearest neighbors within the pseudo-correspondence set $\{m(\boldsymbol{p}_i^{(t-1)})\}$. Let these $k$ nearest pseudo-neighbors be $\{m(\boldsymbol{p}_{m_1}^{(t-1)}), \ldots, m(\boldsymbol{p}_{m_k}^{(t-1)})\}$. Since each pseudo-correspondence point $m(\boldsymbol{p}_m^{(t-1)})$ corresponds to a source point $\boldsymbol{p}_m^{(t-1)}$ from the previous frame, it can inherit the corresponding world coordinate $\hat{\boldsymbol{p}}_m^{(t-1)}$. Next, we compute the prior world coordinate $\tilde{\boldsymbol{p}}_j^{(t)}$ for the actual current-frame point $\boldsymbol{p}_j^{(t)}$ by performing a distance-weighted average of the world coordinates carried by these $k$ pseudo-neighbors:

$$\tilde{\boldsymbol{p}}_j^{(t)} = \sum_{l=1}^{k} w_{j,l} \cdot \hat{\boldsymbol{p}}_{m_l}^{(t-1)}, \tag{9}$$

the weight is defined as $w_{j,l} = \frac{1/d_{j,l}}{\sum_{s=1}^{k} 1/d_{j,s}}$, where $d_{j,l}$ represents the Euclidean distance from the actual point $\boldsymbol{p}_j^{(t)}$ to its $l$-th pseudo-neighbor $m(\boldsymbol{p}_{m_l}^{(t-1)})$.

**Table 1: Average translation error(m) and rotation error(°) on the QE-Oxford dataset. The best results are indicated in bold, and the second-best results are underlined.**

| | Category | Methods | 15-13-06-37 | 17-13-26-39 | 17-14-03-00 | 18-14-14-42 | Average[m/°] |
|---|---|---|---|---|---|---|---|
| **QE-Oxford** | APR | PointLoc (2022 Sens.J) [44] | 10.75/2.36 | 11.07/2.21 | 11.53/1.92 | 9.82/2.07 | 10.79/2.14 |
| | | PosePN++ (2022 PR) [57] | 4.54/1.83 | 6.44/1.78 | 4.89/1.55 | 4.64/1.61 | 5.13/1.69 |
| | | PoseSOE (2022 PR) [57] | 4.17/1.76 | 6.16/1.81 | 5.42/1.87 | 4.16/1.70 | 4.98/1.79 |
| | | STCLoc (2022 TITS) [56] | 5.14/1.27 | 6.12/1.21 | 5.32/1.08 | 4.76/1.19 | 5.34/1.19 |
| | | NIDALoc (2023 TITS) [55] | 3.71/1.50 | 5.40/1.40 | 3.94/1.30 | 4.08/1.30 | 4.28/1.38 |
| | | HypLiLoc (2023 CVPR) [43] | 5.03/1.46 | 4.31/1.43 | 3.61/1.11 | 2.61/1.09 | 3.89/1.27 |
| | SCR | SGLoc (2023 CVPR) [23] | 1.79/1.67 | 1.81/1.76 | 1.33/1.59 | 1.19/1.39 | 1.53/1.60 |
| | | LiSA (2024 CVPR) [51] | 0.94/1.10 | 1.17/1.21 | 0.84/1.15 | 0.85/1.11 | 0.95/1.14 |
| | | LightLoc (2025 CVPR) [21] | 0.82/1.12 | 0.85/1.07 | 0.81/1.11 | 0.82/1.16 | 0.83/1.12 |
| | | RALoc (2025 ICCV) [52] | 1.52/1.28 | 1.71/1.27 | 1.37/1.20 | 1.41/1.23 | 1.51/1.24 |
| | | **TempLoc (Ours)** | **0.74/0.99** | **0.77/0.99** | **0.60/0.95** | **0.77/1.05** | **0.72/0.99** |

***Loss Function.*** The propagated prior coordinates $\tilde{\mathbf{P}}^{(t)}$ and their associated uncertainties are jointly supervised by the loss $\mathcal{L}_{PCG}$, which is analogous to the definitions in Equations (2)-(4).

### 3.3 Uncertainty-Guided Coordinate Fusion

For each LiDAR scan, we can obtain two groups of world coordinates: (1) the *measurement* estimation $(\hat{\mathbf{P}}^{(t)}, \hat{u}^{(t)})$ comprising the coordinates and uncertainties directly predicted by the Global Coordinate Estimation, and (2) the *prior* estimation $(\tilde{\mathbf{P}}^{(t)}, \tilde{u}^{(t)})$ obtained from the Prior Coordinate Generation. To obtain the final and more accurate coordinate estimation, we design the Uncertainty-Guided Coordinate Fusion module, as shown in Figure 4.

Inspired by the Kalman filtering [15, 47], which involves an adaptive weighted fusion based on the uncertainty of different sources, we design a data-driven and end-to-end differentiable fusion mechanism. Specifically, for each point, we concatenate its prior uncertainty $\tilde{u}_i^{(t)}$ and measurement uncertainty $\hat{u}_i^{(t)}$ and use a Softmax function to dynamically compute the fusion weights. The Softmax function ensures the weights are positive and sum to one, which provides a probabilistic interpretation for the fusion, and allowing the network to automatically learn how to balance the two sources based on data:

$$[\alpha_i, \beta_i] = \text{Softmax}([-\tilde{u}_i^{(t)}, -\hat{u}_i^{(t)}]), \tag{10}$$

where $\alpha_i$ is the weight assigned to the prior coordinate $\tilde{\boldsymbol{p}}_i^{(t)}$ and $\beta_i$ is the weight for the measurement coordinate $\hat{\boldsymbol{p}}_i^{(t)}$. The final fused coordinates $\bar{\boldsymbol{p}}^{(t)}$ and uncertainty $\bar{u}^{(t)}$ are then computed by weighted summation:

$$\bar{\boldsymbol{p}}_i^{(t)} = \alpha_i \tilde{\boldsymbol{p}}_i^{(t)} + \beta_i \hat{\boldsymbol{p}}_i^{(t)}, \tag{11}$$

$$\bar{u}_i^{(t)} = \alpha_i \tilde{u}_i^{(t)} + \beta_i \hat{u}_i^{(t)}. \tag{12}$$

The weighted outputs are applied to both the coordinates and their associated uncertainty scores, resulting in the fused estimations with consistence and reliability.

***Loss Function.*** During the training phase, the fused results are supervised by the loss $\mathcal{L}_{Fuse}$, which is analogous to the losses in Equations (2)-(4). Finally, the overall loss of our framework can be formulated as:

$$\mathcal{L}_{full} = \lambda_1 \mathcal{L}_{GCE} + \lambda_2 \mathcal{L}_{PCG} + \lambda_3 \mathcal{L}_{Fuse}, \tag{13}$$

where $\lambda_1$, $\lambda_2$, and $\lambda_3$ are set to 0.3, 0.3, and 0.4, respectively, and control the weights of the three losses.

## 4 Experiments

In this section, we first describe the experimental setup, including the benchmark datasets, training details, and baseline methods (Sec. 4.1). We then conduct extensive experiments to compare the proposed TempLoc with the state-of-the-art methods on typical benchmarks (Sec. 4.2). Lastly, a series of ablation studies are conducted (Sec. 4.3).

### 4.1 Experimental Setup

***Benchmark Datasets.*** We conduct evaluation experiments on two widely-used benchmark datasets for outdoor LiDAR-based localization: the **Oxford** RobotCar Dataset [4] and the **NCLT** Dataset [6]. During evaluation, unified localization metrics are adopted: average translation error(m) and average rotation error(°).

**Oxford RobotCar Dataset.** This dataset was collected in central Oxford, UK, using a Velodyne-32 LiDAR sensor, with GPS/INS ground truth. The route length is 10km, covering an area of about 2.5km² of urban road environments. Consistent with other localization methods, we use four routes for training (11-14-02-26, 14-12-05-52, 14-14-48-55,18-15-20-12) and four routes for testing (15-13-06-37,

**Table 2: Average translation error(m) and rotation error (°) on the Oxford dataset. The best results are indicated in bold, and the second-best results are underlined.**

| | Category | Methods | 15-13-06-37 | 17-13-26-39 | 17-14-03-00 | 18-14-14-42 | Average[m/°]↓ |
|---|---|---|---|---|---|---|---|
| Oxford | APR | PointLoc (2022 Sen.J) [44] | 12.42/2.26 | 13.14/2.50 | 12.91/1.92 | 11.31/1.98 | 12.45/2.17 |
| | | PosePN++ (2022 PR) [57] | 9.59/1.92 | 10.66/1.92 | 9.01/1.51 | 8.44/1.71 | 9.43/1.77 |
| | | PoseSOE (2022 PR) [57] | 7.59/1.94 | 10.39/2.08 | 9.21/2.12 | 7.27/1.87 | 8.62/2.00 |
| | | STCLoc (2022 TITS) [56] | 6.93/1.48 | 7.55/1.23 | 7.44/1.24 | 6.13/1.15 | 7.01/1.28 |
| | | NIDALoc (2023 TITS) [55] | 5.45/1.40 | 7.63/1.56 | 6.68/1.26 | 4.80/1.18 | 6.14/1.35 |
| | | HypLiLoc (2023 CVPR) [43] | 6.88/1.09 | 6.79/1.29 | 5.82/**0.97** | 3.45/**0.84** | 5.74/**1.05** |
| | SCR | SGLoc (2023 CVPR) [23] | 3.01/1.91 | 4.07/2.07 | 3.37/1.89 | 2.12/1.66 | 3.14/1.88 |
| | | LiSA (2024 CVPR) [51] | 2.36/1.29 | 3.47/1.43 | 3.19/1.34 | **1.95**/1.23 | 2.74/1.32 |
| | | LightLoc (2025 CVPR) [21] | 2.33/1.21 | 3.19/1.34 | 3.11/1.24 | 2.05/1.20 | 2.67/1.25 |
| | | RALoc (2025 ICCV) [52] | 3.19/4.10 | 3.87/3.96 | 3.32/3.87 | 2.59/3.71 | 3.24/3.91 |
| | | **TempLoc (Ours)** | **2.22/1.05** | **3.08/1.12** | **2.94**/1.06 | 1.99/1.07 | **2.55**/1.07 |

**Table 3: Average translation error(m) and rotation error(°) on the NCLT dataset. The best results are indicated in bold, and the second-best results are underlined. † denotes the removal of certain erroneous test segments following the LightLoc, details are provided in Appendix Sec. 8.1.**

| | Category | Methods | 2012-02-12 | 2012-02-19 | 2012-03-31 | 2012-05-26† | Average[m/°]↓ |
|---|---|---|---|---|---|---|---|
| NCLT | APR | PointLoc (2022 Sens.J) [44] | 7.23/4.88 | 6.31/3.89 | 6.71/4.32 | 9.55/5.21 | 7.45/4.58 |
| | | PosePN++ (2022 PR)[57] | 4.97/3.75 | 3.68/2.65 | 4.35/3.38 | 8.42/4.30 | 5.36/3.52 |
| | | PoseSOE (2022 PR)[57] | 13.09/8.05 | 6.16/4.51 | 5.24/4.56 | 13.27/7.85 | 9.44/6.24 |
| | | STCLoc (2022 TITS) [56] | 4.91/4.34 | 3.25/3.10 | 3.75/4.04 | 7.53/4.95 | 4.86/4.11 |
| | | NIDALoc (2023 TITS) [55] | 4.48/3.59 | 3.14/2.52 | 3.67/3.46 | 6.32/4.67 | 4.40/3.56 |
| | | HypLiLoc (2023 CVPR) [43] | 1.71/3.56 | 1.68/2.69 | 1.52/2.90 | 2.29/3.34 | 1.80/3.12 |
| | SCR | SGLoc (2023 CVPR) [23] | 1.20/3.08 | 1.20/3.05 | 1.12/3.28 | 3.48/4.43 | 1.75/3.46 |
| | | LiSA (2024 CVPR) [51] | 0.97/**2.23** | 0.91/**2.09** | 0.87/**2.21** | 3.11/**2.72** | 1.47/**2.31** |
| | | LightLoc (2025 CVPR) [21] | 0.98/2.76 | 0.89/2.51 | 0.86/2.67 | 3.10/3.26 | 1.46/2.80 |
| | | RALoc (2025 ICCV) [52] | 1.61/4.71 | 1.61/4.88 | 1.58/4.71 | 3.51/5.40 | 2.07/4.92 |
| | | **TempLoc (Ours)** | **0.74**/2.48 | **0.74**/2.22 | **0.68**/2.35 | **2.04**/3.20 | **1.05**/2.56 |

17-13-26-39, 17-14-03-00, 18-14-14-42). QE-Oxford [23] represents an enhanced version of the original Oxford Radar RobotCar dataset, where ground-truth poses are refined and corrected through trajectory alignment techniques, thereby yielding more accurate and reliable experimental results. We discuss this in detail in **Appendix Sec. 8.2**.

**NCLT Dataset.** This dataset was collected at the University of Michigan's North Campus using a Segway robotic platform equipped with a Velodyne-32 LiDAR sensor. It encompasses both indoor and outdoor environments with seasonal variations. Ground truth is obtained using GPS refined by SLAM techniques. The average route length is approximately 5.5km. Following SGLoc [23], we select four routes for training (2012-01-22, 2012-02-02, 2012-02-18, 2012-05-11) and four routes for testing (2012-02-12, 2012-02-19, 2012-03-31, 2012-05-26).

***Training Details.*** The proposed TempLoc method is implemented with PyTorch[31]. To ensure fairness, the comparison methods utilize the officially provided code and pre-trained models. During training, the batch size is set to 64, and the Adam[18, 46] optimizer is used with an initial learning rate of 0.001. For the Oxford dataset, the point cloud is sampled with a voxel size of 0.25, while for the NCLT dataset, the voxel size is set to 0.3. The parameter $\tau$ is related to learning rate decay, we reduce $\tau$ (Equation (1)) by ×0.7 every 6 epochs. All experiments are conducted on the platform with Intel Xeon CPU@2.30GHZ with two NVIDIA RTX 3090Ti GPUs. **Further details are in the supplementary material.**

***Baselines.*** The proposed TempLoc method is compared with several state-of-the-arts, including both APR and SCR methods. Specifically, the APR baselines contain single-frame methods such as PointLoc [44], PosePN++ [57], PoseSOE [57], and HypLiLoc [43], as well as temporal approaches like STCLoc [56] and NIDALoc [55]. For SCR methods, we consider SGLoc [23], LiSA [51], LightLoc [21], and RALoc [52]. To verify the advantages of our map-free method over retrieval-based localization, we also compare our approach with BEVplace++ [25].

## 4.2 Evaluation

***Evaluation on the Oxford Dataset.*** We first evaluate TempLoc on the **QE-Oxford** dataset. As shown in Table 1, TempLoc outperforms existing methods in average accuracy. Compared with the baseline LightLoc, it reduces translation and rotation errors from $0.83m$ / 1.12° to $0.72m$ / 0.99°. Against temporal-constraint-based methods STCLoc and NIDALoc, TempLoc improves translation error from $4.28m$ to $0.72m$ and orientation error from 1.38° to 0.99° using only two consecutive frames, demonstrating substantial gains with minimal temporal dependency.

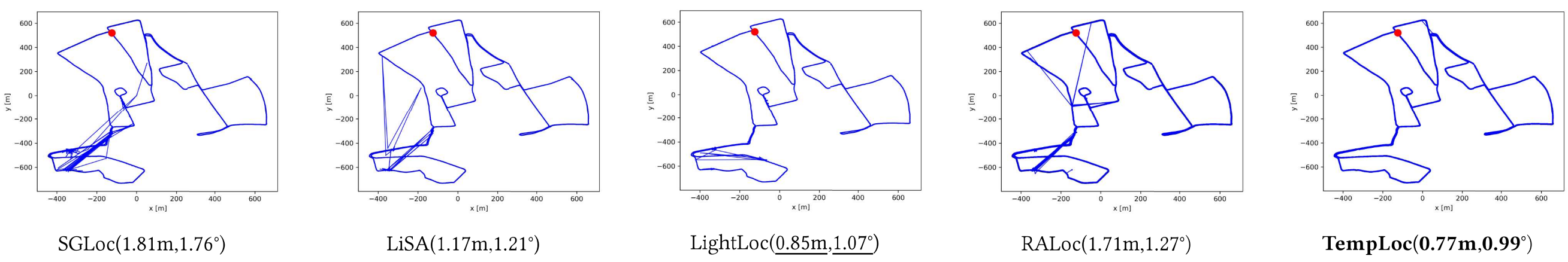

**Figure 5: Visualization of localization results on the QE-Oxford dataset. The blue trajectories represent the predicted results, while black lines denote the ground truth. The best results are indicated in bold, and the second-best results are underlined.**

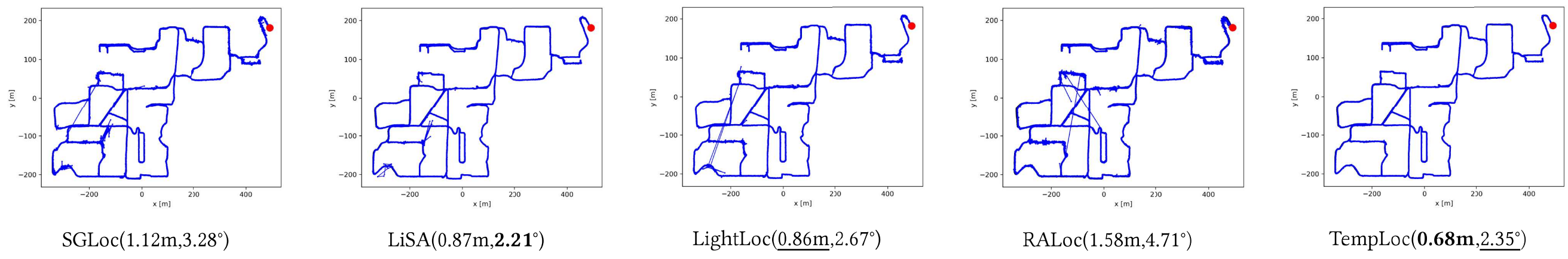

**Figure 6: Visualization of localization results on the NCLT dataset. The blue trajectories represent the predicted results, while black lines denote the ground truth. The best results are indicated in bold, and the second-best results are underlined.**

**Table 4: Real-time, Overhead and Performance on NCLT.**

| Method | Real-time (↓) | Test GPU (↓) | Storage (↓) Model+Map | Error(↓) [m/°] | Recall@1 <5m(↓) |
|---|---|---|---|---|---|
| SGLoc(23'CVPR) [23] | 75ms | 5.2GB | 425MB+0MB | 3.48/4.43 | 95.2% |
| LightLoc(25'CVPR) [21] | **48ms** | 2.1GB | 70MB+0MB | 3.10/3.26 | 96.5% |
| BEVplace++(25'TRO) [25] | 54ms | **1.95GB** | 17MB+841MB | 11.83/4.64 | 92.7% |
| **TempLoc (Ours)** | 68ms | 2.3GB | 77MB+0MB | **2.04/3.20** | **98.3%** |

**Table 5: Localization Performance in Dynamic Scenes on the QE-Oxford Dataset (translation/rotation error in [m/°]) ↓.**

| Method | Scene1 | Scene2 | Scene3 | Scene4 |
|---|---|---|---|---|
| SGLoc (2023 CVPR) [23] | 2.31/1.23 | 3.15/1.56 | 1.77/1.06 | 7.61/2.05 |
| LiSA (2024 CVPR) [51] | 1.97/1.43 | 1.87/1.33 | 1.44/1.01 | 2.07/1.65 |
| LightLoc (2025 CVPR) [21] | 2.20/1.28 | 1.09/0.92 | 0.69/0.98 | 1.77/1.15 |
| RALoc (2025 ICCV) [52] | 1.93/1.34 | 1.00/1.04 | 1.48/1.22 | 3.47/1.78 |
| **TempLoc (Ours)** | **1.00/0.97** | **0.44/0.63** | **0.50/0.54** | **0.40/0.34** |

Figure 5 shows the localization results on the QE-Oxford dataset using trajectory 17-13-26-39. SGLoc, LiSA, RALoc, and LightLoc all suffer from noticeable jumps. In contrast, TempLoc produces a smooth trajectory without erroneous jumps, highlighting stronger robustness via temporally-aware feature fusion.

Due to QE-Oxford's adoption of global trajectory alignment from SGLoc [23], the ground truth error in the **Oxford** dataset was corrected. To ensure experimental fairness, we re-conducted experiments on the Oxford localization baseline, as shown in Table 2. While all methods experienced a decline in accuracy, TempLoc consistently achieves the best performance.

***Evaluation on the NCLT Dataset.*** We further evaluate TempLoc on the NCLT dataset. As shown in Table 3, TempLoc outperforms the second-best method LightLoc with a 28% translation improvement ($1.46m \rightarrow 1.05m$). Although its rotation accuracy is 2.56°, lower than LiSA (2.31°), TempLoc still outperforms LiSA by 29% in translation accuracy ($1.47m \rightarrow 1.05m$). Note that LiSA relies on a heavy semantic segmentation model [20] during training, yet its gain is far inferior to our temporal-aware fusion. Figure 6 visualizes the 2012-03-31 trajectory. SGLoc, LiSA, LightLoc, and RALoc exhibit severe jumps, whereas TempLoc yields a smooth trajectory with top accuracy (0.68m), demonstrating robust performance in complex campus environments.

***Real-time, Overhead and Performance.*** We further compare TempLoc with BEVplace++ [25] and other map-free baselines on NCLT. TempLoc reduces feature storage to 0 MB compared to 841 MB for BEVplace++, and maintains a latency of 68 ms within the 100 ms real-time limit using a 2.3 GB GPU footprint. Crucially, TempLoc outperforms BEVplace++, reducing translation error from $11.83m$ to $2.04m$ and achieving 98.3% recall@1 within $5m$, demonstrating a robust trade-off between efficiency and localization accuracy.

***Localization Analysis under Dynamic Interference.*** As shown in Table 5, we evaluate localization performance on the QE-Oxford sequences 17-14-03-00, which contain multiple dynamic vehicle scenarios. SGLoc, LightLoc, and RALoc suffer from significant performance degradation in the presence of moving vehicles, whereas TempLoc remains unaffected and maintains stable accuracy. This demonstrates the strong robustness of TempLoc in dynamic environments. We discuss this in detail in **Appendix Sec.9**.

***Visualization Comparison and Analysis.*** As shown in Figure 7, we present a qualitative comparison on the QE-Oxford and NCLT datasets. Closer alignment between the predicted and ground-truth point clouds indicates higher localization accuracy. For the QE-Oxford dataset (the left of Figure 7), we select a challenging traffic intersection scene containing numerous dynamic vehicles. These moving objects introduce severe interference to the localization models, resulting in noticeable ghosting and misalignment in the predicted point clouds of all competing methods when compared with the ground truth, whereas TempLoc achieves near-perfect overlap with ground truth. In the densely vegetated NCLT scene (the right of Figure 7), TempLoc leverages its uncertainty estimation module to retain high-certainty structural points, outperforming baseline in point-cloud alignment.

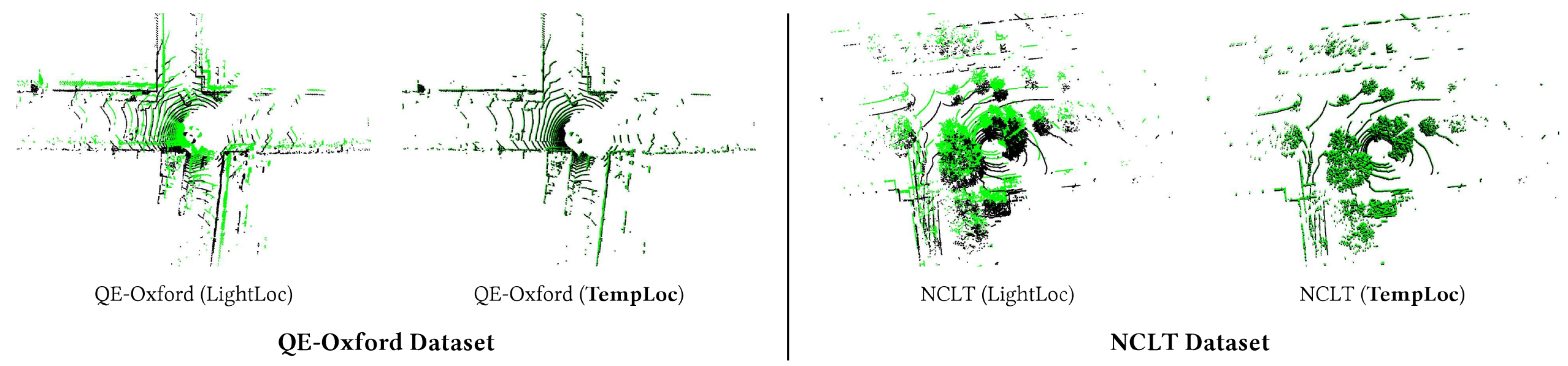

**Figure 7: Localization Visualization Comparison. Experiments were conducted on the QE-Oxford dataset and NCLT datasets. The ground-truth point clouds are depicted in black, while the point clouds at the predicted localization are rendered in green.**

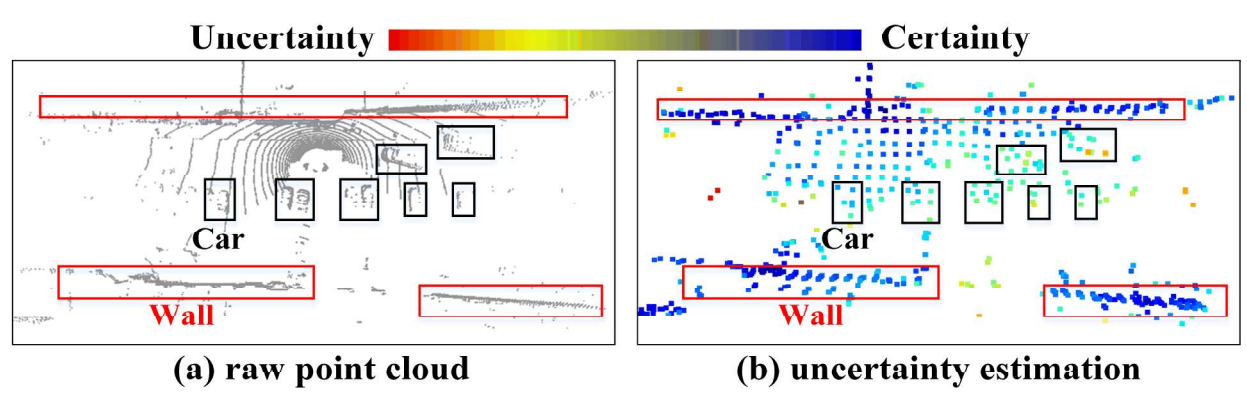

**Figure 8: Uncertainty estimation visualization in dynamic environments. We present the top-down view of the point cloud, with dynamic vehicles highlighted using black bounding boxes.**

**Table 6: Ablation study on different modules. UE: Uncertainty Estimation, PCG+UCF: Prior Coordinate Generation and Uncertainty-guided Fusion. (translation/rotation error in [m/°]) ↓.**

| UE | PCG+UCF | Oxford | QE-Oxford | NCLT |
|---|---|---|---|---|
| | | 2.67m/1.25° | 0.83m/1.12° | 1.46m/2.80° |
| ✓ | | 2.56m/1.09° | 0.74m/1.00° | 1.30m/2.41° |
| ✓ | ✓ | 2.55m/1.07° | 0.72m/0.99° | 1.05m/2.56° |

## 4.3 Ablation Study

***Uncertainty Estimation Visualization.*** To verify that TempLoc can learn more robust features against dynamic objects through temporal awareness and uncertainty estimation, we visualize the point clouds under dynamic scenes in Figure 8. Specifically, Vehicles in the raw point cloud are highlighted with black bounding boxes, and static structures such as buildings have black borders (Figure 8(a)). Figure 8(b) shows that stable structures like buildings exhibit high certainty, while dynamic vehicles show high uncertainty. These results indicate that the network effectively suppresses dynamic object interference, improving localization robustness.

***Ablation Study on Different Modules.*** To validate the effectiveness of the uncertainty estimation (UE) module and the fusion module (PCG+UCF) in TempLoc, we conduct ablation experiments on the Oxford, QE-Oxford, and NCLT datasets, as shown in Table 6. Starting from the baseline, incorporating the uncertainty estimation module (UE) improves translation and rotation accuracy by 4.1%/10.8%/10.9% and 12.8%/10.7%/13.9% across the three datasets, respectively. The relatively modest translation improvement on Oxford is attributed to inherent ground-truth noise in that benchmark. The PCG+UCF module yields smaller gains on Oxford and QE-Oxford, as these datasets contain more stable structures. However, in the complex campus environment of NCLT, PCG+UCF significantly reduces translation error compared with the UE-only variant, boosting overall localization performance by 19% (from 1.30$m$ to 1.05$m$). The ablation results demonstrate that each component plays a critical role in TempLoc.

## 5 Conclusion

In this paper, we present TempLoc, a novel LiDAR relocalization framework designed to overcome the robustness limitations of traditional single-frame methods in challenging dynamic outdoor environments. By explicitly modeling sequential consistency across consecutive LiDAR scans, TempLoc follows a measurement-prediction-fusion paradigm for robust 6-DoF pose estimation without dense map storage or pre-built databases. Extensive experiments conducted on the Oxford, QE-Oxford, and NCLT benchmarks demonstrate that TempLoc consistently outperforms state-of-the-art methods in both translation and rotation accuracy. Crucially, TempLoc transitions from relying solely on instantaneous noisy observations to integrating historical and current information, enhancing robustness in complex outdoor scenarios with dynamic objects.

## Acknowledgments

This work was partially supported by the National Natural Science Foundation of China (No. 62501502).